\documentclass{article}

\PassOptionsToPackage{numbers, compress}{natbib}

\usepackage[preprint]{neurips_2020}
\usepackage[utf8]{inputenc} 
\usepackage[T1]{fontenc}    
\usepackage{hyperref}       
\usepackage{url}            
\usepackage{booktabs}       
\usepackage{amsfonts}       
\usepackage{nicefrac}       
\usepackage{microtype}      
\usepackage{courier}

\usepackage{hyphenat}
\usepackage{times}
\usepackage{epsfig}
\usepackage{graphicx}
\usepackage{amsmath}
\usepackage{amssymb}
\usepackage{microtype}
\usepackage{graphicx}
\usepackage{subfigure}
\usepackage{booktabs} 
\usepackage{relsize}

\usepackage{amsmath}
\usepackage{amssymb}
\usepackage{graphicx}  
\usepackage{subfigure}
\usepackage{enumitem}
\usepackage{graphbox}
\usepackage{amssymb}
\usepackage{bbm}
\usepackage{bm}
\usepackage{amsmath}
\usepackage{comment}
\usepackage{wrapfig}
\usepackage{booktabs}
\usepackage{multirow}
\usepackage{wrapfig}
\usepackage{todonotes}

\definecolor{forestgreen}{RGB}{17, 205, 47}

\newcommand{\etal}{\emph{et al.}}

\definecolor{Mycolor2}{HTML}{006400}

\usepackage{amsmath,amsfonts,bm}

\def\eqref#1{equation~\ref{#1}}

\def\1{\bm{1}}

\DeclareMathAlphabet{\mathsfit}{\encodingdefault}{\sfdefault}{m}{sl}
\SetMathAlphabet{\mathsfit}{bold}{\encodingdefault}{\sfdefault}{bx}{n}

\newcommand{\R}{\mathbb{R}}

\author{Saneem A. Chemmengath\footnotemark[1]$^{\,\,,1}$, Soumava Paul\thanks{Contributed equally. ${}^\dagger$Work done while at IBM Research. Email: \texttt{saneem.cg@in.ibm.com}.}$^{\,\,,\dagger,2}$, Samarth Bharadwaj$^{1}$, \\
\textbf{Suranjana Samanta}$^{1}$, \textbf{Karthik Sankaranarayanan}$^{1}$ \\
$^{1}$ IBM Research, $^{2}$ IIT Kharagpur
}

\begin{document}

\title{Addressing Target Shift in Zero-shot Learning\\ using Grouped Adversarial Learning}





\maketitle

\newtheorem{theorem}{Theorem}

\begin{abstract}

Zero-shot learning (ZSL) algorithms typically work by exploiting attribute correlations to be able to make predictions in unseen classes. However, these correlations do not remain intact at test time in most practical settings and the resulting change in these correlations lead to adverse effects on zero-shot learning performance. In this paper, we present a new paradigm for ZSL that: (i) utilizes the class-attribute mapping of unseen classes to estimate the change in target distribution (\emph{target shift}), and (ii) propose a novel technique called \emph{grouped Adversarial Learning (gAL)} to reduce negative effects of this shift. Our approach is widely applicable for several existing ZSL algorithms, including those with implicit attribute predictions. We apply the proposed technique (\emph{g}AL) on three popular ZSL algorithms: ALE, SJE, and DEVISE, and show performance improvements on 4 popular ZSL datasets: AwA2, aPY, CUB and SUN. We obtain SOTA results on SUN and aPY datasets and achieve comparable results on AwA2.
\end{abstract}










\section{Introduction}

Zero-shot learning (ZSL) algorithms are designed to train classifiers using examples of \emph{seen} classes to be able to generalize and predict \emph{any} set of unseen classes \cite{DAP, zsl_2009}. Such models generalize by utilizing additional information, specifically, semantically relevant mid-level \emph{attributes} that (are assumed to) persist between seen and unseen classes.
Hence, the performance of a ZSL model is governed by its ability to predict these persistent attributes in instances of unseen classes. The standard view of ZSL assumes class-attribute mapping for the test classes is available only at inference time. On the other hand, the transductive ZSL represents a relaxed view \cite{ transductive_2015,song2018transductive} that allows for unlabelled test set as unsupervised additional information. However, obtaining a significant number of instances from unseen classes of interest is not always feasible. 

 \begin{wrapfigure}[15]{R}{0.45\textwidth}
 \begin{center}
 \includegraphics[width=\linewidth]{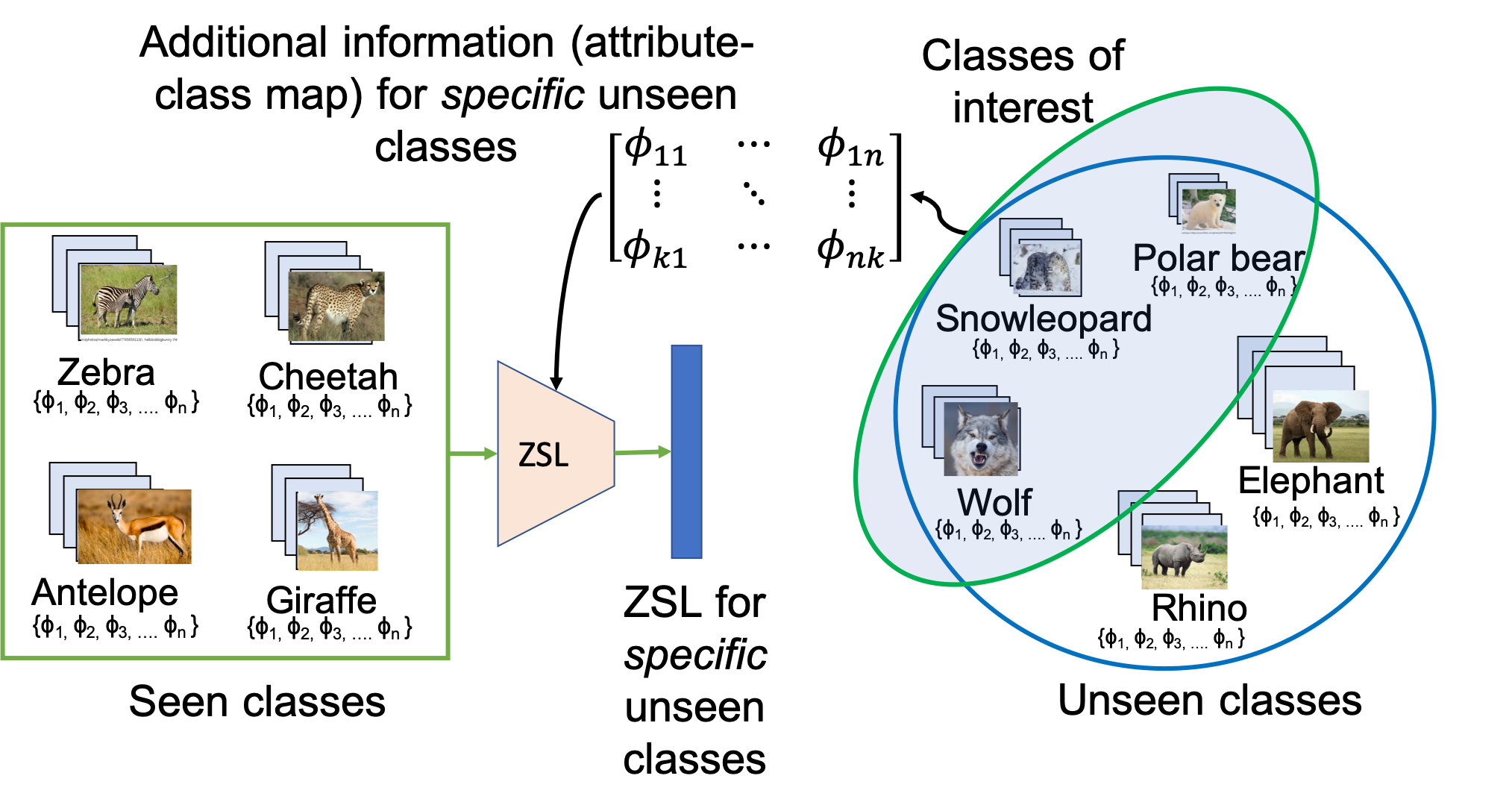}
 \caption{Our approach to Zero shot learning uses attribute class map for the specific unseen classes to minimize \emph{target shift}.}
 \end{center}
 \label{fig:first_example}
\end{wrapfigure}

In ZSL, attribute correlations are useful when the expected label correlation of unseen classes remain consistent with that of train classes. However, we observed that a key reason for the practical difficulty of predicting attributes from instances of unseen classes is the adverse effect of those attribute correlations that are highly likely to change in the test set, we term this effect \emph{correlation shift}. When the attribute predictors of ZSL are viewed as an instance of multilabel classification, the change in the attribute distribution may be viewed with the lens of \emph{domain adaptation} literature as \emph{target shift}~\cite {target_shift_lin02}. 
However, existing target shift correction techniques from domain adaptation use \emph{importance reweighting}, which is not applicable to ZSL (see detail in Sec.\ref{sec:prob_setup}), the shift in correlation between the attributes can be considered as one aspect of target shift. We hypothesize that it is necessary to estimate correlation among attributes in test set to correct correlation shift. We propose to use class-attribute vectors of test classes to estimate test correlation.

In the low-resource scenario of ZSL, it is pragmatic to leverage the more readily available additional information about the attribute space. It is much easier to construct a class-attribute mapping of test classes by utilizing class descriptions from auxiliary sources such as knowledge bases (e.g. Wikipedia). For example, to train a ZSL image classifier for the rare and endangered \emph{Red Wolf} animal, it would be easier to find attributes describing it such as \{slender\hyp{}legged, large, carnivorous, long\hyp{}ears\} from common sources rather than obtaining several samples of \emph{Red Wolf} images.

To the best of our knowledge, this is the first work which addresses the phenomenon of \emph{correlation shift} (as an aspect of target shift) in zero shot learning. The contributions of this work are as follows: (i) As illustrated in Fig.\ref{fig:first_example}, we present a new zero-shot learning paradigm where the classifier can be tailored to a \emph{specific} set of unseen classes by only utilizing additional information such as attribute-class mapping. Specifically, we show that the proposed framework is effective in curtailing \emph{correlation shift} (as an aspect of target shift) between attributes of seen and unseen classes. 
(ii) Building on a principled analysis on a controlled synthetic dataset, we propose \emph{grouped adversarial learning} (\emph{g}AL) paradigm for correlation shift that is universally applicable to any attribute-prediction based ZSL architecture that is end-to-end trainable. We demonstrate performance improvements with \emph{g}AL with three popular ZSL algorithms: ALE~\cite{ALE}, DEVISE~\cite{DEVISE} and SJE~\cite{SJE} on four standard zero-shot learning benchmarks, namely, Animals-with-Attributes-2 (AwA2) \cite{ZSL_survey}, Attribute Pascal and Yahoo (aPY) \cite{apy}, Scene UNderstanding (SUN) \cite{SUNdataset}, and Caltech UCSD Birds (CUB) \cite{WahCUB_200_2011} datasets. (iii) Finally, we release a new experimental benchmark (train-test split) that maximizes correlation-shift between the seen and unseen classes to amplify the problem of correlation shift.


\section{Related Work}\label{sec:related_work}


\textbf{Zero Shot Learning}:
Zero shot learning has been extensively studied in recent years \citep{ZSL_survey, eszsl, SAE, cmtZSL, SYNC, CONSE, tzeng2017adversarial, PSR}. Existing methods in ZSL can be broken down into the following categories : i) intermediate attribute classifiers \citep{DAP}, ii) bilinear compatibility frameworks that treat zero-shot recognition as a ranking problem \citep{ALE, SJE, DEVISE}, iii) linear closed-form solutions optimized by a ridge regression or mean-squared error objective \citep{eszsl, SAE}, iv) non-linear compatibility frameworks \cite{PSR, LATEM, CMT}, v) hybrid models \citep{verma2017simple, CONSE, SYNC, SSE}, and vi) generative models \citep{xian2018feature, Sariyildiz_2019_CVPR, huang2019generative, kumar2018generalized, chen2018zero} based on GANs\cite{goodfellow_gan} or VAE\cite{kingma2013auto} that synthesize images for unseen classes during training. Xian \etal \cite{ZSL_survey} performed an extensive benchmarking of several such algorithms under a common benchmark protocol, representation vectors and hyper-parameter tuning, and showed that the performance of linear compatibility models are comparable with the more complex joint representation-based hybrid models. In a slightly different line of work, some approaches \citep{zhang2017learning, lei2015predicting, shigeto2015ridge} propose techniques to tackle the now well-known \emph{hubness} problem in ZSL, created by projecting seen and unseen class image features to the attribute (semantic) space. Besides inductive and conventional ZSL, there exists an extensive line of work on transductive \citep{song2018transductive, fu2014transductive, fu2015transductive} and generalized ZSL \citep{chao2016empirical, kumar2018generalized, liu2018generalized, schonfeld2019generalized, huang2019generative} as well. However, such approaches are not the focus of this work.


\textbf{Target shift}:
Previous literature on target shift \cite{da_target_shift, target_shift_lin02, da_label_shift} utilize importance re-weighting over training instances to match the probability of train set with that of test set. This process performs poorly when the cardinality of label set is large (curse of dimensionality). This setting also assumes that instances of labels in test set should strictly be a subset of that of train set (see \emph{Sec}.\ref{sec:main_idea}). This  is not the case in zero shot learning, where different label (attribute) combinations define a class, and train and test sets have different groups of classes. 

\textbf{Label correlation}:
Addressing the negative effects of label correlations has been previously explored in the areas of machine learning under various terms: 
debiasing \cite{debias_mitigating, debias_controllable_invariance, debias_fair}, privacy preservation literature \cite{debias_minimax, debias_privacy, debias_censoring}, and multi-task learning \cite{mtl_esclusive_lasso, mtl_unrelated_tasks, jayaraman2014decorrelating}. 
De-biasing and privacy preservation settings are interested in protected variables or sensitive/private variables that are correlated with the desired label. 
In multitask learning (MTL), several regularization based methods are proposed to mitigate negative effects of label correlation 
\cite{mtl_esclusive_lasso, jayaraman2014decorrelating, mtl_unrelated_tasks} which attempt to decorrelate label predictors using special regularizers that enforce predictors of different labels to use non-overlapping set of features. The overall intent of these techniques is to decorrelate a multi label classification model. However, such regularizers are not applicable for learned features with end-to-end trainable neural networks. 

\section{Proposed Framework} 

\subsection{Problem Formulation} \label{sec:prob_setup} 
\textbf{Notations and problem setup for ZSL}: Given a seen dataset $\mathcal{D}^s = \{(x_i^s, y_i^s)\}_{i=1}^N$ of $N$ points where $x_i^s \in \mathcal{X}$ denotes the instance and $y_i^s$ denotes class label from seen classes $y_i^s \in \mathcal{Y}^s$. For the ZSL problem setup, the aim is to build a model, which trained on $\mathcal{D}^s$, can classify instances of unseen classes $x_i^u$ with labels $y_i^u \in \mathcal{Y}^u$, where $\mathcal{Y}^s$ and $\mathcal{Y}^u$ are disjoint. 
Apart from instances and class labels, for every class $y \in \mathcal{Y}^s \cup \mathcal{Y}^u$, we are provided with $D$ dimensional class-attribute vector $\phi^y \in \{0,1\}^D$, where $\phi^y_m = 1$ if $m$-th attribute is present in class $y$, otherwise $0$. Attribute vectors connect seen and unseen classes in the semantic space that aids in inference during test time. We use $\Phi^s = \{\phi^y\}_{y \in \mathcal{Y}^s}$ to denote set of class-attribute vectors of seen classes and $\Phi^u$ to denote that of unseen classes. Note that we use train with seen and test with unseen interchangeably in this paper.

\textbf{Attribute target shift} : In this work, we focus on those ZSL algorithms that map input instances to attributes either explicitly or implicitly. Given an input instance ($x$), an explicit model predicts binary attribute vector ($\widehat{\phi}(x)$) whereas implicit methods provide soft scores for each attribute ($\widehat{\phi}(x) \in \mathbb{R}^D$). $c$ is predicted as the class for an instance if attribute vector  $\phi^c$ is most \emph{compatible} with predicted attribute vector $\widehat{\phi}(x)$.
Emphasizing only on the task of predicting attributes of instances, we view ZSL as a special case of transfer learning for multilabel classification where the attribute distributions ($P_{\phi}$) differ from seen to unseen classes. 
We view the change in the attribute distributions ($P_{\phi}$) as domain adaptation under \textit{target shift} \cite{da_label_shift, target_shift_change_in_y, target_shift_lin02}, where attribute marginals for the training set (seen classes $P^{s}_{\phi}$) and that for test set (unseen classes $P^{u}_\phi$) are different
while, conditionals $P_{X|\phi}$ remain the same. 
Since correcting for target shift requires $P^{u}_\phi$ along with the training data, we use set of attribute vectors of unseen classes $\Phi^u$ to estimate $P^{u}_\phi$ by assuming that all unseen classes are equally likely in the test set.
We could also estimate $P^{u}_\phi$ from unlabelled test data using Black Box Shift Estimation (BBSE)~\cite{da_label_shift}, however, obtaining unlabelled test instances changes the problem setting to transductive-ZSL, which is beyond the current scope.

Existing approaches to correct target shift, such as importance re-weighting~\cite{cost_sensitive, da_target_shift}, match attribute distributions of train and test set by appropriately weighing each instance by $P^u_{\phi}/P^s_{\phi}$ in the loss function. However, importance re-weighting can't be extended to ZSL since attribute vector $\phi$ in train set do not appear in the test set essentially letting all the weights be zero ($P^{u}_\phi = 0$ for all $P^{s}_\phi > 0$). 


\subsection{Adversarial learning to address Target Shift}
\label{sec:main_idea}

We begin the description of our approach to correcting target shift in multilabel case with a two-label problem. We start here in order to systematically build the arguments and merits of our design choices that we later extend to more labels and ultimately to ZSL. 
We begin with a standard \emph{feature extractor} $h: \mathcal{X} \rightarrow \mathbb{R}^d$, which projects instance $x$ to a latent feature vector $h(x)$. These features are then mapped to labels space, in the case of the two label problem, as $\phi_1$ using a attribute predictor $f_1$, and $\phi_2$ using $f_2$. Note, $\phi_1$, $\phi_2$ predictions for $x$ are $f_1(h(x))$ and $f_2(h(x))$, respectively. Let the two-attribute distributions be given by $p(\phi_1, \phi_2)$, that can be \emph{factorized} into three constituents: the marginals ($p(\phi_1)$ and $p(\phi_2)$), and the correlation coefficient ($\rho_{\phi_1, \phi_2}$) between $\phi_1$ and $\phi_2$. Hence, target shift for the two attributes can be viewed as the combination of shifts in two marginal distributions and a further shift in correlation among attributes. We later refer to the portion of change attributed to correlation as \emph{correlation shift}, which we propose to correct with adversarial learning.

We adopt the popular formulation of adversarial learning designed for unsupervised domain adaptation \cite{ganin2016domain} and widely used to debias models \cite{debias_privacy, debias_minimax, debias_mitigating}. Specifically, for prediction model of $\phi_1$, we use $\phi_2$ as an adversarial task and vice versa ($\phi_2$ against $\phi_1$)., i.e., separate models are used to predict each attribute. If $\phi_1$ and $\phi_2$ are correlated in the train set but relatively uncorrelated in the test set, the objective is to identify a feature extractor for $\phi_1$ that is disinclined to utilize feature information pertaining to $\phi_2$, thereby ensuring $\widehat{\phi}_1$ and $\phi_2$ remain uncorrelated, hence correcting \emph{correlation shift}. The above intuition is grounded in the objective function: 
\begin{align}
\label{eq:adv2}
    \min_{f_1, h} \max_{f_2} ~~ \sum_{i=1}^N~ \ell(~f_1(h(x_i)),~ \phi_1(x_i)) - \lambda ~ \ell(~f_2(h(x_i)),~\phi_2(x_i)),
\end{align}
where, $\ell(\cdot,\cdot)$ is binary classification loss and $\lambda \in \mathbb{R}^+$ is the adversarial weight, the hyperparameter which controls the trade-off between predicting $\phi_1$, and decorrelating $(\hat{\phi_1},~\phi_2)$. Intuitively one can see that in Eq.\ref{eq:adv2}, higher the value of $\lambda$, lesser the information to predict $\phi_2$ would be present in $h(x)$, resulting in lower correlation between predicted attribute $\widehat{\phi}_1$ and $\phi_2$. A similar model for predicting $\phi_2$ with $\phi_1$ as adversarial arm will be used. 

The primary advantage of adversarial learning in correcting correlation shift in ZSL over re-weighting methods, is that it can be applied to ZSL methods with implicitly predicted attributes. Further, with the right weighting scheme, predictors for single attribute may have several adversarial branches connected to it that simultaneously minimize all pairwise correlation shift against it. We use gradient reversal layer with SGD to optimize the objective as done in \cite{ganin2015unsupervised}. Choosing the right $\lambda$ is essential to correcting target shift. We show that having an estimate of correlation shift helps in finding better $\lambda$ values using some heuristics (Sec \ref{sec: adv_weight}). 

\subsubsection{Synthetic experiments}
\label{sec:synth_expts}

We continue to systematically study the two-label problem and the effects of adversarial training to curtail target shift. We now generate synthetic data as it allows us to create training and test sets with specific feature correlations which is not otherwise possible on real data. This analysis reveals some counter-intuitive observations that motivate the proposed formulation which is presented later in Sec.\ref{sec:gal}.

\begin{figure}[!ht]
\centering
\subfigure[]{\includegraphics[align=c, width=0.14\linewidth]{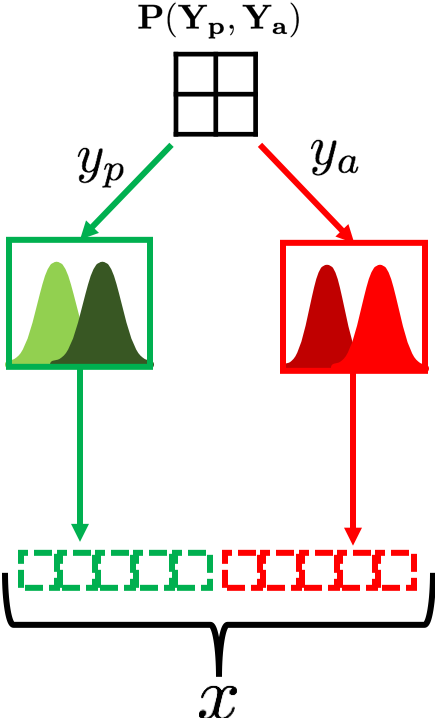}}\hfill
\subfigure[]{\includegraphics[align=c, width=0.4
\linewidth]{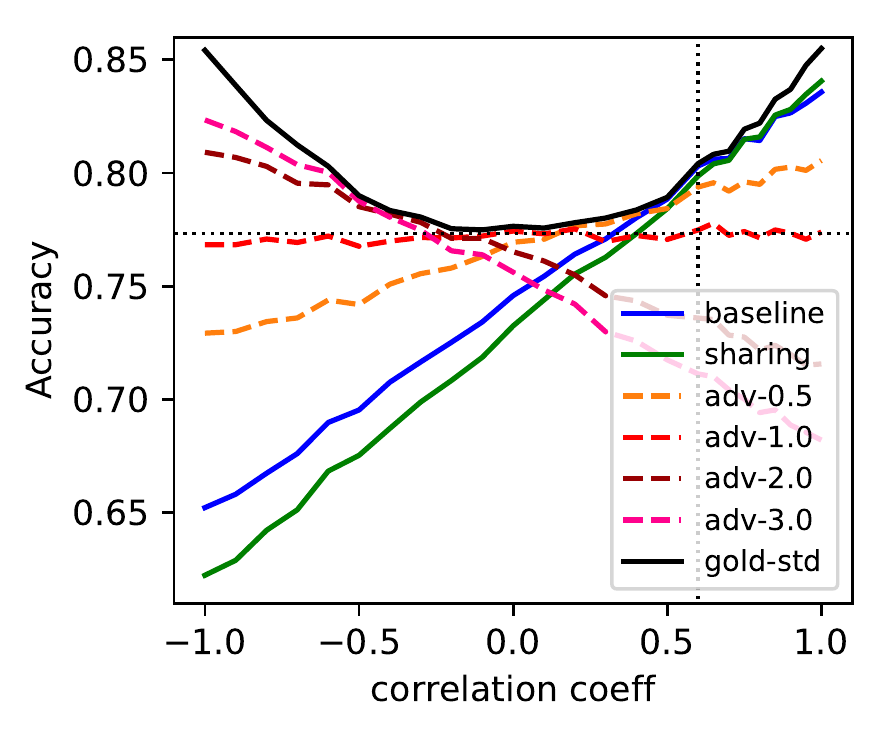}}\hfill
\subfigure[]{\includegraphics[align=c, width=0.35\linewidth]{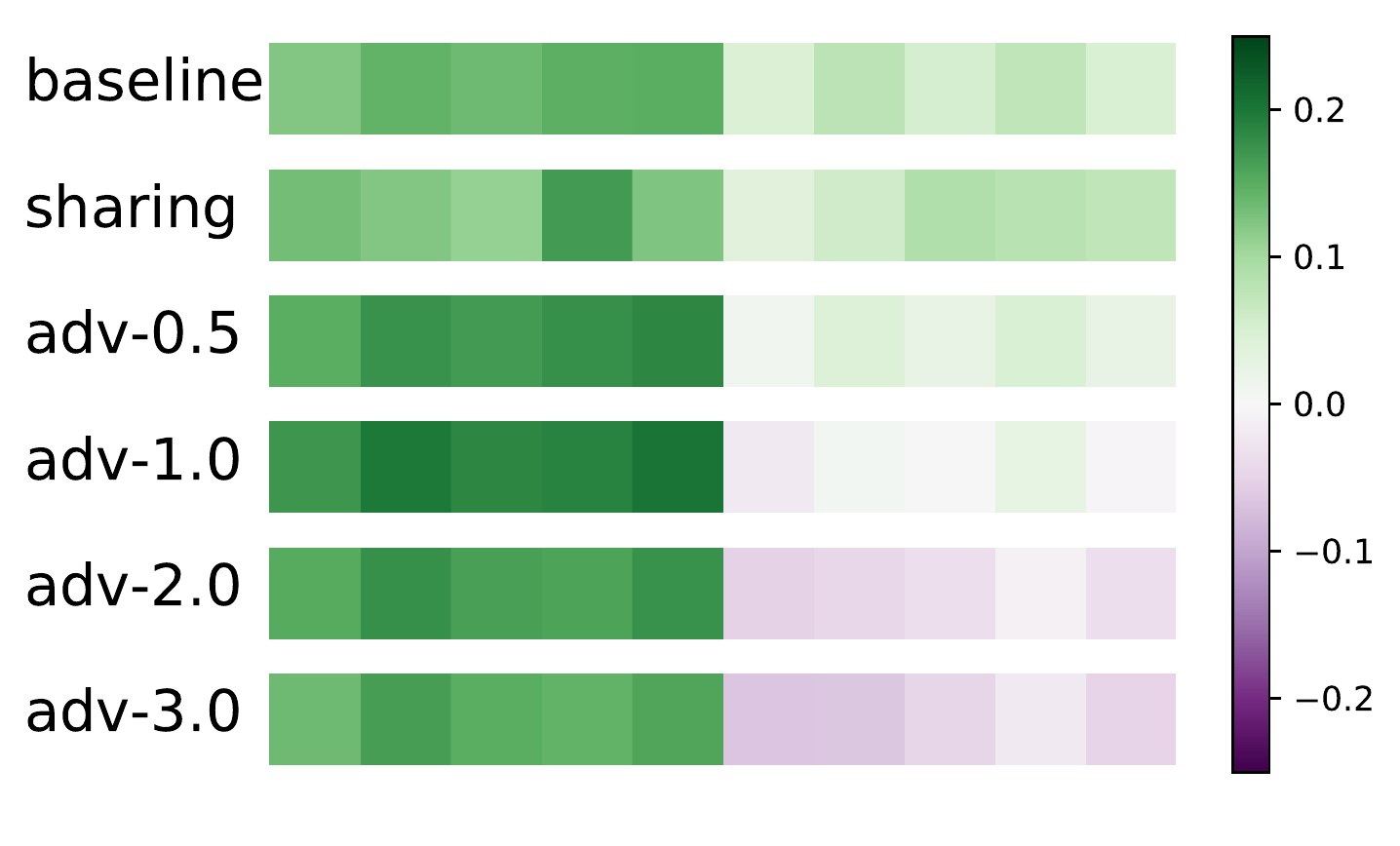}}
\caption{Synthetic experiments analysis: (a) Probabilistic data generation to create data with target shift (b) Model accuracy on test set with varying label correlation, when model was trained on training set with correlation $0.6$ (vertical dotted lines in the diagram) (c) Model weights on features. Note that best models for $y_p$ and $y_a$ give equal positive weights to first five and last five features respectively.}
\label{fig:synthetic_result}
\end{figure}

\textbf{Data Generation}\footnote{details and reproducible python notebook in supplementary.}: The synthetic dataset consists of real vectors $x \in \R^{10}$, with corresponding binary labels $y_p$ and $y_a$ (primary and auxiliary). As show in Fig.\ref{fig:synthetic_result}(a) we generate data from a probabilistic generative system with \emph{different} label distributions $P(Y_p,Y_a)$ for train and test sets, with same conditional $P(X|Y_p,Y_a)$ throughout, thereby creating a target shift between them. A data point $x$ is generated by first sampling $(y_p,y_a)$ from the label distribution. Then the features are sampled from two 5 dimensional multivariate Gaussian distributions with identity covariance matrix such that $\mathcal{N}_{k=5}$($\mu_1$,$\mathbb{I}$), \textit{if} $y_p$=$1$ or \textit{else}, $\mathcal{N}_{k=5}$($\mu_2$,$\mathbb{I}$), 
where $\mu_1, \mu_2$ are chosen such that the best linear classifier has positive and equal weights for all the 5 features for both $y_p$ and $y_a$, therby ensuring all 5 features are equally important. Further, we have $P(Y_p$=$1)$ = $P(Y_a$=$1)$=$0.5$ to ensure no class-imbalance exists between the two labels. The distance between the Gaussian distributions  corresponding to primary label and auxiliary label is fixed at $1.5$, which corresponds to Bayes accuracy of $77.3\%$. We fix label correlation in train set to $0.6$ and create test sets with correlations from $-1$ to $1$. We aim to analyze the predictive power for the primary label $y_p$ trained at a given label correlation and evaluated against multiple test sets with varying label correlations. Specifically, we train the models on training set with $P(Y_p|Y_a) = 0.8$ and test performance on test sets which only differ from the train set in $P(Y_p|Y_a)$. We sample 1000 instances for train and a very high number of 50,000 instances in test to avoid sampling bias in all evaluations.

\noindent We compare following algorithms in this analysis: A \textbf{Baseline} linear logistic regression classifier trained only on the primary label $y_p$, a \textbf{Sharing} model with two-label MLP and one hidden layer (of two neurons) that predict both $y_p$ and $y_a$. Here, the common hidden layer encourages sharing between modes, and \textbf{Adv-$\lambda$}, which is an adversarial learning model with one hidden layer of two neurons (as encoder), a label predictor for primary label $y_p$ and a discriminator to predict auxiliary label $y_a$ with an adversarial weight $\lambda$. All the models are linear functions with no activation functions. 

\noindent\textbf{Observations and Insights}: Fig.\ref{fig:synthetic_result}(b) illustrates the test accuracy on primary label prediction against all label correlations in test set. The performance of \textit{baseline} model is monotonically affected by the change in correlation between $Y_p$ and $Y_a$. Further, we observe that the performance is less affected when the correlation increases with the same polarity. A similar observation was made by \cite{hendricks2018women} in \textit{bias} setting and is termed bias amplification. On the other hand, adversarial models ($adv$-$1.0$) are more invariant to various label correlations in the test set that is consequence of target shift.
The choice of adversarial weight (hyper-parameter $\lambda$) is critical to the performance of the model for a given test correlation. For instance, in this setup, $\lambda$=$1.0$ is the best choice when test set is uncorrelated i.e., $\rho_{p,a}^{te}=0.0$, whereas a larger $\lambda$ is more suitable for test correlations near $-0.5$. Interestingly, the choice of $\lambda$ even causes the models to achieve \emph{higher} accuracy in a target shifted test set than the training set. Fig.\ref{fig:synthetic_result}(c) visualizes model weights on 10-dimensional feature vector for all models. 
As the $\lambda$ for adversarial models increases, we observe that the model weights for the features corresponding to the auxiliary label are reduced. Furthermore, for larger value of $\lambda$, the model assigns negative weights on features corresponding to $y_a$. Negative weights on last five features imply that the model has captured opposite correlation between labels even though such a correlation is not observed in training. 


\subsection{Grouped Adversarial Learning (\emph{g}AL)}
 \label{sec:gal}
We now describe our novel grouped adversarial learning to correct the effects of target shift in attribute prediction of zero-shot learning algorithms where, typically, a large number of attributes (e.g., parts of animals or birds) are predicted for unseen classes. To reiterate, our framework leverages additional information available about the unseen classes to diminish the effects of correlation shift in their attribute predictors. However, to simply extend the aforementioned intuition requires applying adversarial learning to a large number of attributes, leading to multiple adversarial branches. 
To ensure tractability, we devise a measure termed $\Delta_{corr}$ to weight the adversarial arms. Further, inspired from multi task learning \cite{mtl_clsutering_thrun, mtl_cluster, mtl_whom_to_share, jayaraman2014decorrelating}, we take a course-grained approach and split the attributes into groups such that only inter-group correlation shift is minimized. 
Our approach is suitable to several ZSL algorithms that produce scores corresponding to attributes. In this work, we specifically apply \emph{g}AL to three popular ZSL methods: ALE~\cite{ALE}, DEVISE~\cite{DEVISE}, and SJE~\cite{SJE}. 

\subsubsection{Attribute importance with \protect{{$\Delta_{corr}$}}}
\label{sec: adv_weight}

For attributes $\phi_1$ and $\phi_2$, we estimate correlation coefficient for seen classes $\rho_s(\phi_1,\phi_2)$ from labelled train set and that of test set $\rho_u(\phi_1, \phi_2)$ using class-attribute mapping. $\Delta_{corr}$ is defined as:
\begin{align}
    \label{eq:dcorr}
    \Delta_{corr}(\phi_1,\phi_2) = \max\{~\text{sgn}(\rho_s(\phi_1,\phi_2))~(\rho_s(\phi_1,\phi_2) - \rho_u(\phi_1, \phi_2)), 0\}.
\end{align}

We showed in Sec. \ref{sec:synth_expts} and Eq.~\ref{eq:adv2} that higher adversarial weight is necessary to counteract a large correlation shift. However, when there is higher correlation in test set than that in train set (with same sign), we see that adversarial learning degrades the performance. Hence, we propose an adversarial weighting scheme using $\Delta_{corr}$ such that attribute pairs with positive $\Delta_{corr}$ are permitted to be adversarial to each other with $\lambda \times \Delta_{corr}$ as adversarial weights, where $\lambda$ is the common hyperparameter across all pairs of attributes.


\subsubsection{Attribute Grouping}
\label{sec:adv_cl}
For a given attribute predictor, we propose to retain only attributes from outside its group as adversarial branches thereby permiting the predictors of attributes of same group to \emph{share} feature representation and leverage their correlations. Earlier works rely on group memberships that are based on semantic similarity of attributes~\cite{jayaraman2014decorrelating} or human perceptions. However, in the context of target shift, we hypothesize that grouping tasks based on correlation shift may be more beneficial. Specifically, the proposed measure of correlation shift, $\Delta_{corr}$, should be low among attribute pairs in the same group and high across groups. To achieve this, we form groups by clustering attributes using spectral co-clustering \cite{co_clustering} with $\Delta_{corr}$ as the distance measure. Nevertheless, we also report our results on semantic groups (whenever applicable) for a fair comparison. 

\subsubsection{Model Architecture}
Given group memberships of attributes and the weighting scheme, we propose a one-vs-all architecture for label prediction, with every group jointly predicting the member attributes constrained by all other groups as adversarial branches.
Let $L$ denote number of groups attributes were split into. In the model, first we have feature extractors $h_1, h_2, \ldots,h_L$, which projects input instances $x$ to $L$ latent representations,
each corresponding to a group. 
Further, to each feature extractor $h_i$, we connect one \textbf{primary branch} $f_{ii}$ which maps to attributes of group $i$ and $(L-1)$ \textbf{adversarial branches} $f_{ij}:j\neq i$ which maps to attributes of group $j$. 
$f_{ij}(h_i(x)) \in \mathbb{R}^{d_j}$, provides scores for each attribute in group $j$. 
So, a model with $L$ groups would have a total of $L$ primary arms and $L(L-1)$ adversarial arms. 
The primary arm of the group latent representation is responsible for predicting all the group attributes, thus enabling sharing. During backpropagation, each latent representation is updated from the primary arm and adversarially updated from the remaining $(L-1)$ adversarial arms. 
The objective function for \emph{g}AL is, \vspace{-5pt}
\begin{align}
\label{eq:gal}
    \min_{f_{ii}, h_i} \max\limits_{\substack{f_{ij}\\ i\neq j}} ~~ 
    \mathlarger{\mathlarger{\sum}}_{k=1}^N~
    \Bigg[~
    &\ell_{ZSL}([f_{11}(h_i(x_k)), \ldots, f_{LL}(h_L(x_k))], \Phi^s, y_k)\nonumber\\
    &- \lambda ~\sum_{i=1}^L \sum_{\substack{j=1\\ j\neq i}}^L ~\Delta_{ij}~~ \ell_{adv}(f_{ij}(h_i(x_k)), \phi_j(x_k)) \Bigg],
\end{align}
where $\ell_{ZSL}$ is the loss function of any ZSL method which takes in score vector on attributes (given in the equation as concatenation of group of attributes)\footnote{$[\cdot,\ldots,\cdot]$ denotes concatenation of vectors.}
and set of class-attribute vectors $\Phi$ to predict class label.
$\Delta_{ij}$ is the fixed adversarial weight between groups $i$ and $j$ which is the highest pairwise $\Delta_{corr}$ between members of group $i$ and $j$ computed using Eq.\ref{eq:dcorr},
$\lambda$ is the hyperparameter to control overall trade-off between class prediction and correcting correlation shift, and $\phi_j(x_k)$ is attribute vector of group $j$ for instance $x_k$. $\ell_{adv}$ is a multilabel classification loss.

We can apply the gAL technique on any ZSL algorithm whose loss functions takes scores over attributes as input. 
We apply gAL on three popular ZSL methods in our experiments: ALE~\cite{ALE}, DEVISE~\cite{DEVISE} and SJE~\cite{SJE}. 
In all these three methods, class score is the dot product of class-attribute vector and attribute scores (this is called linear compatibility in \cite{ZSL_survey}). Score for class $c$ is computed as $\hat{y}_c = [f_{11}(h_i(x_k)), \ldots, f_{LL}(h_L(x_k))]^{\top}\phi^c$. 
Given class prediction vector $\hat{y}(x)$ and ground truth $y(x)$, one could apply any multiclass classification loss here. DEVISE uses SVM-rank based loss, while ALE and SJE uses some extra weighting schemes over the SVM-Rank loss.
We tried a fourth ZSL method of using a categorical cross-entropy loss over the class predictions denoted as \texttt{softmax}~\cite{xian2018feature}.

To optimize gAL objective function, special gradient flipping layer before the adversarial arms called \textit{gradient reversal layer} \cite{ganin2015unsupervised} is used. This ensures that the model performs poorly in prediction of adversarial labels in each group, leading to \emph{decorrelated} learning of attributes. For the attribute predictors in adversarial branches, there could be effects of class imbalance from the target shift, hence we choose $\ell_{adv}$ as balanced binary cross-entropy (bce) loss.

\section{Experiments}
\label{sec:exp}

\subsection{Datasets and Protocol} 

\noindent \textbf{Protocol}: We follow the experimental protocol introduced in previous literature \cite{ZSL_survey} for the four datasets described in Table \ref{tab:dataset}. The experimental protocol is designed such that the validation set is also zero-shot in nature. We utilize the 2048-D ResNet-101 \cite{he2016deep} feature representation and ``attribute-class prior" matrices provided by the authors of \cite{ZSL_survey}. 

\begin{table}[!ht] \scriptsize
\vspace{-0.2cm}
\centering
\resizebox{\textwidth}{!}{%
\begin{tabular}{|l|c|c|c|c|c|c|c|}
\hline
\multicolumn{1}{|c|}{\multirow{2}{*}{Dataset}} & \multirow{2}{*}{\#attributes} & \multirow{2}{*}{\begin{tabular}[c]{@{}c@{}}\#seen classes\\ (train + val)\end{tabular}} & \multirow{2}{*}{\begin{tabular}[c]{@{}c@{}}\#unseen \\ classes \end{tabular}} & \multirow{2}{*}{\begin{tabular}[c]{@{}c@{}}\#seen images\\ (train + val)\end{tabular}} & \multirow{2}{*}{\begin{tabular}[c]{@{}c@{}}\#unseen \\ images \end{tabular}} & \multicolumn{2}{c|}{$\Delta$ $\text{corr}$} \\ \cline{7-8} 
\multicolumn{1}{|c|}{} &  &  &  &  &  & mean & \multicolumn{1}{c|}{mean @top 50\%} \\ \hline
\textbf{aPY} \cite{apy} & 64 & 15+5 & 12 & 6086+1329 & 7924 & {\bf 0.073} & {\bf 0.145} \\ 
\textbf{AWA2} \cite{ZSL_survey} & 85 & 27+13 & 10 & 20218+9191 & 7913 & {\bf 0.161} & {\bf 0.319} \\ 
\textbf{CUB} \cite{WahCUB_200_2011} & 312 & 100+50 & 50 & 5875+2946 & 2967 & 0.019 & 0.036 \\ 
\textbf{SUN} \cite{SUNdataset} & 102 & 580+65 & 72 & 11600+1300 & 1440 & 0.016 & 0.033 \\ \hline \hline
\textbf{aPY-CS} & 64 & 15+5 & 12 & 4299+6691 & 4349 & {\bf 0.132} & {\bf 0.246} \\ 
\textbf{AWA2-CS} & 85 & 27+13 & 10 & 22103+10383 & 4836 & {\bf 0.255} & {\bf 0.483} \\ 
\textbf{CUB-CS} & 312 & 100+50 & 50 & 5901+2958 & 2929 & 0.041 & 0.076 \\ 
\textbf{SUN-CS} & 102 & 580+65 & 72 & 11600+1300 & 1440 & 0.074 & 0.136 \\ \hline
\end{tabular}%
}
\caption{\label{tab:dataset} Statistics of datasets with attribute $\Delta_{corr}$ between train and test sets.}
\end{table}

\noindent \textbf{Correlation-shift analysis and new splits}: Table \ref{tab:dataset} also shows the mean difference in correlation, measured by $\Delta_{corr}$ (Eq. \ref{eq:dcorr}) and $\Delta_{corr}$ measured for the top $50\%$ of attribute pairs. We highlight the significantly high change in correlation for the AWA2 and aPY datasets. Further, we generate a new experimental split of train, validation and test through a greedy selection approach, termed \textbf{CS split} (correlation-shift split), such that the difference in correlation (measured by $\Delta_{corr}$) is maximized, while keeping the class-count per split unchanged from the existing protocol \cite{ZSL_survey}. Under these CS splits, $\Delta_{corr}$ for AWA2 and aPY is even higher than before. 
The considerable drop in performance of baselines on these splits further highlights the problems of target shift and showcases the ability of \emph{g}AL to correct for them. We skip experimentation on CUB-CS and SUN-CS as the increase in $\Delta_{corr}$ in not significant.

\begin{wraptable}[24]{R}{0.6\textwidth}
\vspace{-0.5cm}
\scriptsize
\begin{center}
\begin{tabular}{|l|l|l|l|l|l|}
\hline
&\multicolumn{1}{|c|}{\textbf{Method}} & \textbf{aPY} & \textbf{AWA2} & \textbf{CUB} & \textbf{SUN} \\ \hline
\multirow{13}{*}{$\alpha$} & DAP \cite{DAP} & 33.8 & 46.1 & 40.0 & 39.9 \\
& IAP \cite{DAP}& 36.6 & 35.9 & 24.0 & 19.4 \\
& CONSE \cite{CONSE}& 26.9 & 44.5 & 34.3 & 38.8\\
& CMT  \cite{CMT}& 28.0 & 37.9 & 34.6 & 39.9 \\
& SSE \cite{SSE}& 34.0 & 61.0 & 43.9 & 51.5 \\
& LATEM \cite{LATEM}& 35.2 & 55.8 & 49.3 & 55.3 \\
& ESZSL \cite{eszsl}& 38.3 & 58.6 & 53.9 & 54.5 \\
& ALE \cite{ALE}& 39.7 & 62.5 & 54.9 & 58.1 \\
& DEVISE \cite{DEVISE}& 39.8 & 59.7 & 52.0 & 56.5 \\
& SJE \cite{SJE}& 32.9 & 61.9 & 53.9 & 53.7 \\
& SYNC \cite{SYNC}& 23.9 & 46.6 & 55.6 & 56.3 \\
& SAE \cite{SAE}& 8.3 & 54.1 & 33.3 & 40.3 \\ 
& GFZSL \cite{verma2017simple} & 38.4 & \textbf{63.8} & 49.3 & 60.6 \\ \hline
\multirow{4}{*}{$\beta$}& SP-AEN \cite{chen2018zero}& 24.1 & 58.5 & 55.4 & 59.2 \\
& f-CLSWGAN \cite{xian2018feature}& -- & -- & \textbf{61.5} & \textcolor{blue}{62.1} \\
& QFZSL \cite{song2018transductive}& -- & \textcolor{blue}{63.5} & \textcolor{blue}{58.8} & 56.2 \\
& PSR \cite{PSR}& 38.4 & \textbf{63.8} & 56.0 & 61.4 \\ \hline
\multirow{6}{*}{$\gamma$} & ALE* & 32.8 & 52.9 & 50.0 & 61.9 \\
& \textbf{ALE-gAL} & 38.3\textsuperscript{\textcolor{Mycolor2}{$\uparrow$ 5.5}} & 58.2\textsuperscript{\textcolor{Mycolor2}{$\uparrow$5.3}} & 52.3\textsuperscript{\textcolor{Mycolor2}{$\uparrow$2.3}} & \textbf{62.2}\textsuperscript{\textcolor{Mycolor2}{$\uparrow$0.3}} \\
& DEVISE* & 33.3 & 57.7 & 44.1 & 55.7 \\
& \textbf{DeViSE-gAL} & 38.9\textsuperscript{\textcolor{Mycolor2}{$\uparrow$5.6}} & 59.4\textsuperscript{\textcolor{Mycolor2}{$\uparrow$1.7}} & 51.7\textsuperscript{\textcolor{Mycolor2}{$\uparrow$7.6}} & 57.4\textsuperscript{\textcolor{Mycolor2}{$\uparrow$1.7}} \\
& SJE* & 32.9 & 58.3 & 49.4 & 53.5 \\
& \textbf{SJE-gAL} & \textbf{40.5}\textsuperscript{\textcolor{Mycolor2}{$\uparrow$7.6}} & 62.2\textsuperscript{\textcolor{Mycolor2}{$\uparrow$3.9}} &  53.2\textsuperscript{\textcolor{Mycolor2}{$\uparrow$3.8}} & 60.3\textsuperscript{\textcolor{Mycolor2}{$\uparrow$6.8}} \\
& \texttt{softmax} & 33.8 & 55.4 & 50.1 & 61.7 \\
& \textbf{\texttt{softmax}-gAL} & \textcolor{blue}{40.0\textsuperscript{\textcolor{Mycolor2}{$\uparrow$6.2}}} & 62.1\textsuperscript{\textcolor{Mycolor2}{$\uparrow$6.7}} & 52.2\textsuperscript{\textcolor{Mycolor2}{$\uparrow$2.1}} & 60.8\textsuperscript{\textcolor{red}{$\downarrow$0.9}} \\ \hline
\end{tabular}
\caption{\label{tab:comp} ($\alpha$) Performance reported in \cite{ZSL_survey}, ($\beta$) recent approaches following same settings, ($\gamma$) performance improvement with \emph{g}AL on three ZSL algorithms.}
\end{center}
\end{wraptable}

\subsection{Results and Discussion}
The experimental results of \emph{g}AL on the standard benchmark \cite{ZSL_survey} and our novel correlation-shift splits are reported in tables \ref{tab:comp} and \ref{tab:cs_exp} respectively. We report class-averaged top-1 accuracies for all datasets. Highest accuracies for each dataset are shown in \textbf{bold} and second best numbers in \textcolor{blue}{blue}. 

We first show performance of ZSL algorithms reported by \cite{ZSL_survey} in Table \ref{tab:comp}:$\alpha$ for easy reference. Table \ref{tab:comp}:$\beta$ shows other recent methods reported on the same benchmarks. In the absence of available public implementations of ALE~\cite{ALE}, SJE~\cite{SJE} and DeViSE~\cite{DEVISE}, we use a public Python implementation\footnote{\scriptsize All baselines (marked $*$) computed from: \url{https://github.com/mvp18/Popular-ZSL-Algorithms}.} whose performance is shown in Table \ref{tab:comp}:$\gamma$ (marked $*$). Also shown are the corresponding \emph{g}AL variants of these algorithms, built from the same codebase (available in supplementary with detailed instructions). We also include the \texttt{softmax} baseline \cite{xian2018feature} trained with categorical cross-entropy loss. Except for \texttt{softmax}-\emph{g}AL on SUN, we report substantial improvement in performance over baseline for all four datasets. The magnitudes of improvement are indicated in \textcolor{Mycolor2}{green}. The highest improvement was observed for SJE-\emph{g}AL on aPY and DeViSE-\emph{g}AL on CUB, giving a boost of \textbf{7.6\%} over baseline.

\begin{wraptable}[11]{R}{0.4\textwidth}
\vspace{-0.5cm}
\scriptsize
\begin{center}
\begin{tabular}{|l|l|l|l|l|}
\hline
\multicolumn{1}{|c|}{\textbf{Method}}  & \textbf{aPY-CS} & \textbf{AWA2-CS} \\ \hline
 ALE* & 21.1 & 25.3\\
\textbf{ALE-\emph{g}AL} & 24.3\textsuperscript{\textcolor{Mycolor2}{$\uparrow$3.2}} & \textbf{42.5}\textsuperscript{{\textcolor{Mycolor2}{$\uparrow$17.2}}}\\
DEVISE* & 19.5 & 33.1\\
\textbf{DEVISE-\emph{g}AL} & \textbf{25.7}\textsuperscript{\textcolor{Mycolor2}{$\uparrow$6.2}} & 38.2\textsuperscript{{\textcolor{Mycolor2}{$\uparrow$5.1}}}\\ 
SJE* & 18.7 & 27.9\\
\textbf{SJE-\emph{g}AL} & 23.9\textsuperscript{\textcolor{Mycolor2}{$\uparrow$5.2}} & 40.2\textsuperscript{\textcolor{Mycolor2}{$\uparrow$12.3}}\\ 
\texttt{softmax} & 18.4 & 32.1 \\
\textbf{\texttt{softmax}-\emph{g}AL} & \textcolor{blue}{24.6\textsuperscript{\textcolor{Mycolor2}{$\uparrow$6.2}}} & \textcolor{blue}{41.5\textsuperscript{\textcolor{Mycolor2}{$\uparrow$9.4}}} \\ \hline
\end{tabular}
\caption{\label{tab:cs_exp} Performance of \emph{g}AL variants on our proposed \textbf{CS splits}.}
\end{center}
\end{wraptable}

The approaches corrected for correlation shift with \emph{g}AL compare favourably with existing approaches on AWA2, SUN, and aPY datasets, achieving SOTA numbers on aPY ({\bf40.5\%}) with SJE-\emph{g}AL and SUN ({\bf 62.2\%}) with ALE-\emph{g}AL. Further, \emph{g}AL improves SJE on AWA2 by 3.9\% to 62.2\%, marginally lower than SOTA of 63.8\%. 
The failure to achieve SOTA on CUB dataset can be attributed to the relatively low correlation shift and the hard task of predicting large number of attributes (312, largest among the 4 datasets) for class inference. However, \emph{g}AL variants continue to perform better than baselines here also.
 
It is interesting to compare our proposed linear compatibility approach (network of linear layers with regularizers) to a non-linear compatibility based method from Table \ref{tab:comp} such as PSR\cite{PSR} or GAN-based methods like SP-AEN\cite{chen2018zero} and f-CLSWGAN\cite{xian2018feature}, that generate additional data to aid training. Note that QFZSL is a transductive algorithm, and the accuracies reported here correspond to the inductive variant. 

On our newly introduced \textbf{CS splits}, the improvement over baseline is more pronounced as shown in table \ref{tab:cs_exp}. The highest improved is seen for ALE-\emph{g}AL on AWA2-CS of \textbf{17.2\%}. The considerably lower accuracies of all approaches compared to Table \ref{tab:comp} demonstrate the difficulties faced by existing ZSL algorithms in conditions of high correlation shift. Consequently, the significant improvements over baseline shows the effectiveness of \emph{g}AL. 

 
All \emph{g}AL variants presented here are based on groups formed by spectral co-clustering\cite{co_clustering} with $\Delta_{corr}$ as the distance measure (see Sec. \ref{sec:adv_cl}). AWA2 and CUB datasets additionally provide semantic grouping of attributes that have been extensively utilized in previous literature\cite{jayaraman2014decorrelating}. However, we observe that the groups formed by co-clustering provide superior empirical performance (see Appendix Sec. A.1). Further, these groups continue to maintain semantic relevance. For instance, the cluster \{\emph{‘lean’, ‘swims’, ‘fish’, ‘arctic’, ‘coastal’, ‘ocean’, ‘water’}\} clearly represents the aquatic animal classes of AWA2.

As mentioned, the adversarial weighting scheme and the choice of hyperparameter $\lambda$ are critical to \emph{g}AL performance. Relevant ablations and model parameter details are also included in the supplementary material. 
\section{Summary}
This paper shows that our grouped adversarial learning coupled with adversarial weighting strategies can be effective in curtailing target-shift in zero shot learning settings and consequently improving performance. 
\begin{itemize}[leftmargin=*]
    \item Traditional zero-shot learning algorithms utilize a set of seen classes (and associated information such as attributes-class mapping) to prepare a classifier for \emph{any} set of unseen classes. This paper presents a variant of zero-shot learning that utilizes additional information from \emph{specific} unseen classes of attributes-class mapping to create a tailored classifier. We show that such a paradigm of zero shot learning can be useful for correcting \emph{target shift} in attributes.
    \item By utilizing the additional information to design and weight the proposed grouped adversarial learning, we substantially improve the performance of three popular ZSL algorithms on four standard benchmark datasets, reaching SOTA on two of them. 
    \item A functional and flexible PyTorch implementation was built for the experimental evaluation of this work along with extensive hyper-parameter tuning heuristics that are essential in training multiple adversarial arms. It has been included with the supplementary material.
\end{itemize}

\newpage
\section{Broader Impact}
The techniques discussed in this paper broadly advance the training of zero-shot classification models, i.e., learning to classify without examples, by mitigating the effects of target-shift. We analyze the impact of our work on both aspects of machine learning with an assumption that substantial empirical improvements and resources are applied to this line of research well beyond the preliminary evaluation presented in this work.

\subsection{Impact of superior zero-shot learning algorithms}
The paradigm of zero-shot learning explored in this research, if matured to its full potential, equips practitioners to transition an ML model's learning from current to new (but related) entities using only additional descriptive information. The ability to transition to new classes with such data efficiency will have transformational impact to applied areas of machine learning. For instance, a disease prediction model may adapt to a new variant with slightly different pathology using only its definition. However, machine learning for detecting or profiling may also benefit from such a model update, provided learning is done efficiently.    

 \begin{wrapfigure}[16]{R}{0.4\textwidth}
 \centering
 \subfigure[typical sample]{\includegraphics[width=0.45\linewidth,height=40pt]{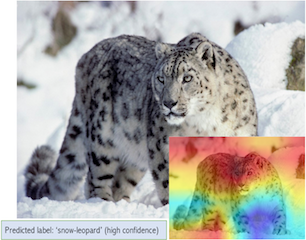}}
 \subfigure[typical sample]{\includegraphics[width=0.45\linewidth,height=40pt]{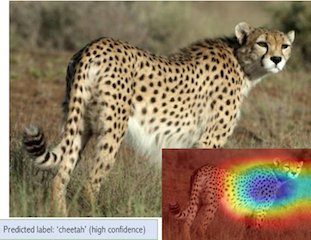}} \\
 \subfigure[rare sample]{\includegraphics[width=0.45\linewidth,height=40pt]{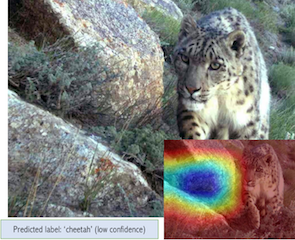}}
 \subfigure[rare sample]{\includegraphics[width=0.45\linewidth,height=40pt]{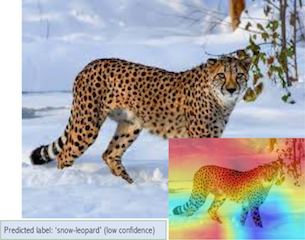}}
 \caption{Visual task of classifying ``cheetah \emph{vs.} snow-leopard''.}
 \label{fig:impact_ex}
 \end{wrapfigure}

\subsection{Impact of panacea solution for target shift}
Similarly, a near perfect solution to target shift can potentially impact generalization, transfer and domain adaptation of ML. Let us consider the visual task of classifying ``cheetah \emph{vs.} snow-leopard'', such as those illustrated in Fig. \ref{fig:impact_ex} -- a task which ideally should primarily focus on the animal's appearance. However, a large portion of these images also contain various secondary/auxiliary cues of the typical habitat of the animals in the background, i.e., tall grass and snow (see Figs \ref{fig:impact_ex} (a) \& (b)) which are, in principle, unrelated to the animal's appearance. An archetypal model is deceived by the co-occurrence of auxiliary cues of habitat over the animal's primary appearance features such as complex fur patterns (see Figs \ref{fig:impact_ex} (c) \& (d)). The poor performance of this visual classifier can be attributed to the \emph{shift} in correlation between appearance and habitat labels between train and test sets. It must be noted that while the formulation of the problem is motivated in zero-shot learning, the corrosive effects of unintended correlations is a disposition of any supervised learning task from simple binary classification to recent popular supervised tasks such as object detection, captioning, or visual dialog. These and other applications of supervised  learning in not only vision, but also other modalities such as speech and text, have substantial impact to our lives. 

The authors readily acknowledge their limitation in foreseeing other impact of this work.

{
\bibliographystyle{plainnat}
\bibliography{main}
}

\onecolumn
\appendix 

\section{Supplementary materials}

\subsection{Ablation}

\noindent \textbf{Results on semantic groups}: As mentioned in Section 3.3, previous literature utilized a semantic grouping of attributes based on human intuition of similarity. However, we find that the cluster groups (groups formed by spectral co-clustering) via $\Delta_{corr}$ provides grouping that best minimize the chance of correlation shift. For a fair comparison, we show experiments on AWA2, CUB, and AWA2-CS performed with \emph{g}AL variants based on semantic groups accompanying the datasets. These are represented by the combinations \textbf{sg+Eq} and \textbf{sg+$\Delta_{corr}$} in Table \ref{tab:gAL_ablations}. To reiterate, AWA2 comes with 10 semantic groups (e.g. nutrition, habitat) for grouping its 85 attributes whereas CUB is provided with 28 semantic groups (e.g. bill shape, wing color) for grouping its 312 attributes. The results show that cluster groups consistently outperform semantically grouped models, where available. On the challenging AWA2-CS protocol, we observe a 7\% difference showcasing the importance of the proposed grouping.

\noindent \textbf{Effect of adversarial weighting scheme}: In Section 3.3, we present a adversarial weighting scheme such that classifier loss from the primary task is given a fixed weight of 1, while all adversarial arms are weighted proportional to $\Delta_{corr}$. Here, we present a comparison where all adversarial arms are equally weighted. These are represented by the combinations \textbf{sg+Eq} and \textbf{cg+Eq} in Table \ref{tab:gAL_ablations}. An improvement of 2.5\% was observed on the AWA2-CS protocol. Further, we observe a relatively stable loss in training.

\begin{table}[!h]
\scriptsize
\centering
\begin{tabular}{|l|l|c|c|c|c|c|c|}
\hline
Method & \multicolumn{1}{|c|}{\textbf{Group+Weights}} & \textbf{aPY} & \textbf{AWA2} & \textbf{CUB} & \textbf{SUN} & \textbf{AWA2-CS} & \textbf{aPY-CS} \\ \hline
\multirow{4}{*}{ALE-\emph{g}AL} & sg+Eq & - & 51.9 & 49.9 & - & 33.6 & - \\
& sg+$\Delta_{corr}$ & - & 54.0 & 51.0 & - & 35.5 & - \\
& cg+Eq & 37.3 & 59.2 & 49.9 & 59.5 & 40.2 & 18.5 \\
& cg+$\Delta_{corr}$ & 38.3 & 58.2 & \textcolor{blue}{52.3} & \textbf{62.2} & \textbf{42.5} & 24.3 \\ \hline
\multirow{4}{*}{DeViSE-\emph{g}AL} & sg+Eq & - & 51.6 & 48.7 & - & 33.2 & - \\
& sg+$\Delta_{corr}$ & - & 54.9 & 50.1 & - & 34.8 & - \\
& cg+Eq & 31.9 & 58.2 & 48.7 & 55.6 & 32.6 & 17.5 \\
& cg+$\Delta_{corr}$ & 38.9 & 59.4 & 51.7 & 57.4 & 38.2 & \textbf{25.7} \\ \hline
\multirow{4}{*}{SJE-\emph{g}AL} & sg+Eq & - & 54.3 & 50.5 & - & 34.0 & - \\
& sg+$\Delta_{corr}$ & - & 54.6 & 51.3 & - & 33.1 & - \\
& cg+Eq & 33.0 & \textbf{62.2} & 51.0 & 56.1 & 38.4 & 10.6 \\
& cg+$\Delta_{corr}$ & \textbf{40.5} & \textbf{62.2} & \textbf{53.2} & 60.3 & 40.2 & 23.9 \\ \hline
\multirow{4}{*}{\texttt{softmax}-\emph{g}AL} & sg+Eq & - & 52.9 & 48.5 & - & 35.5 & - \\
& sg+$\Delta_{corr}$ & - & 54.5 & 49.3 & - & 35.4 & - \\
& cg+Eq & 37.1 & 61.6 & 50.4 & 59.9 & 40.5 & 14.2 \\
& cg+$\Delta_{corr}$ & \textcolor{blue}{40.0} & \textcolor{blue}{62.1} & 52.2 & \textcolor{blue}{60.8} & \textcolor{blue}{41.5} & \textcolor{blue}{24.6} \\ \hline
\end{tabular}
\vspace{0.2cm}
\caption{\label{tab:gAL_ablations} Ablations for all \emph{g}AL variants with best numbers in \textbf{bold}, second best numbers in \textcolor{blue}{blue}. 
Here, sg: semantic groups. cg:cluster groups. Eq: equal weights to all adversarial branches. $\Delta_{corr}$:weights on adversarial branches proportional to $\Delta_{corr}$. Relevant numbers from Tables 2 and 3 are shown again for easy reference.}

\end{table}

\subsection{Effect of adversarial weight $\lambda$}

\begin{wrapfigure}[9]{R}{0.5\textwidth}
\vspace{-1cm}
\subfigure[SJE-\emph{g}AL]{\includegraphics[width=0.45\linewidth]{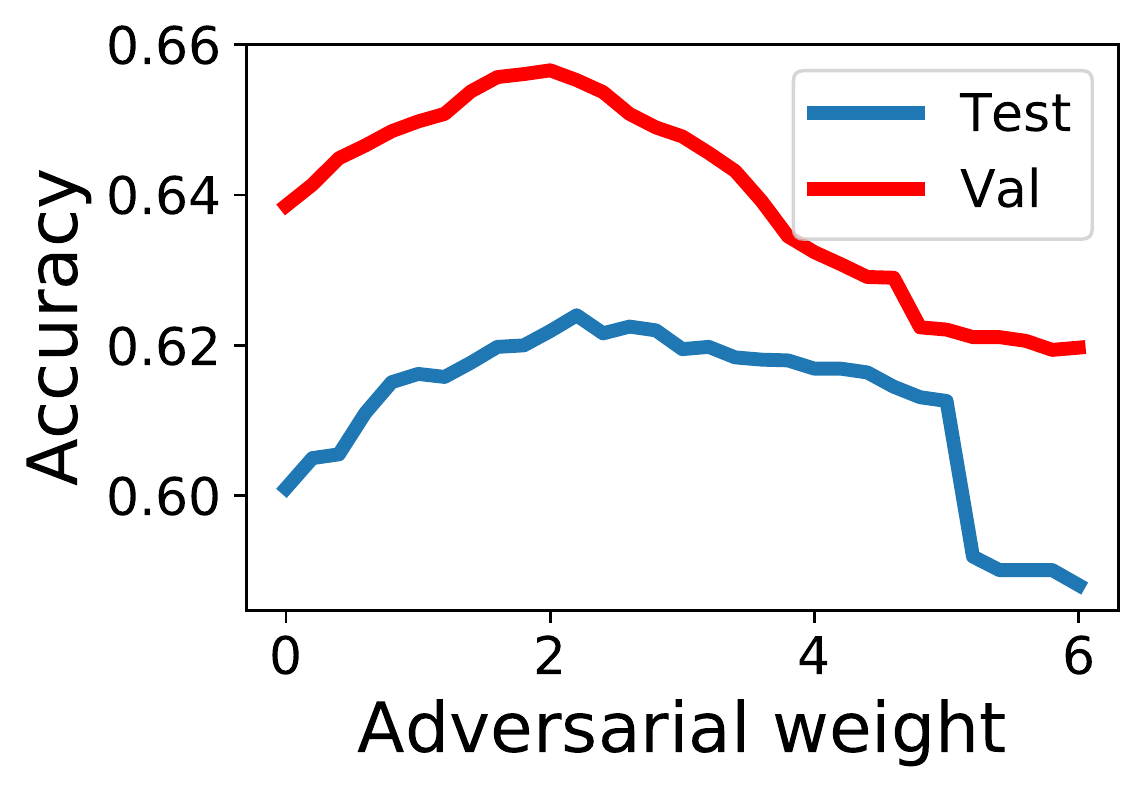}}\hspace{1em}
\subfigure[\texttt{softmax}-\emph{g}AL]{\includegraphics[width=0.45\linewidth]{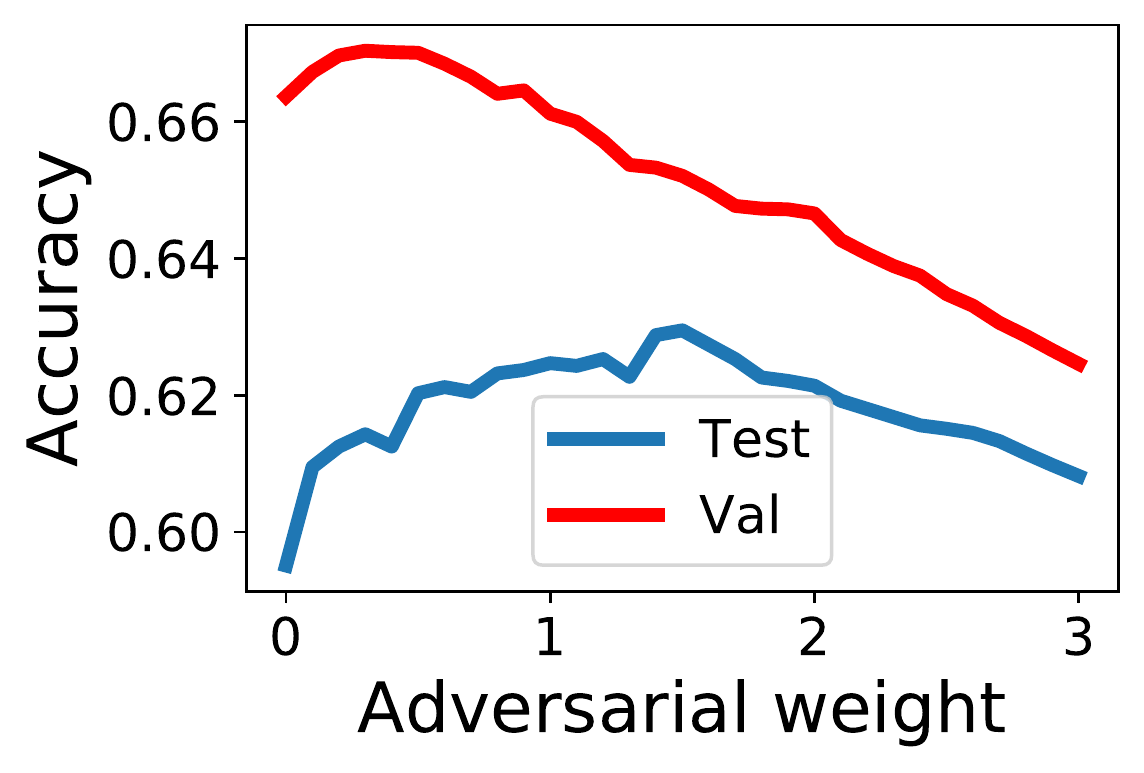}}
\caption{Accuracies on AwA2 dataset with varying adversarial weight ($\lambda$) illustrates the importance of selecting the adversarial weight.}
\label{fig:lambda_plot}
\end{wrapfigure}

Choice of adversarial weight $\lambda$ (Eq.\ref{eq:gal}) is crucial for performance of \emph{g}AL models. In Fig.\ref{fig:lambda_plot}, we observe how the test accuracy rises and drops as adversarial weight increases. This shows the trade-off between predicting classes and correcting correlation shift. 
Best value of adversarial weight is selected using validation accuracy. In Fig.\ref{fig:lambda_plot}(b), the difference in performance of validation and test highlights the difficulty of finding the right value of adversarial weight.

\subsection{Implementation Details}
\begin{figure}[!h]
\centering
\includegraphics[width=0.7\textwidth]{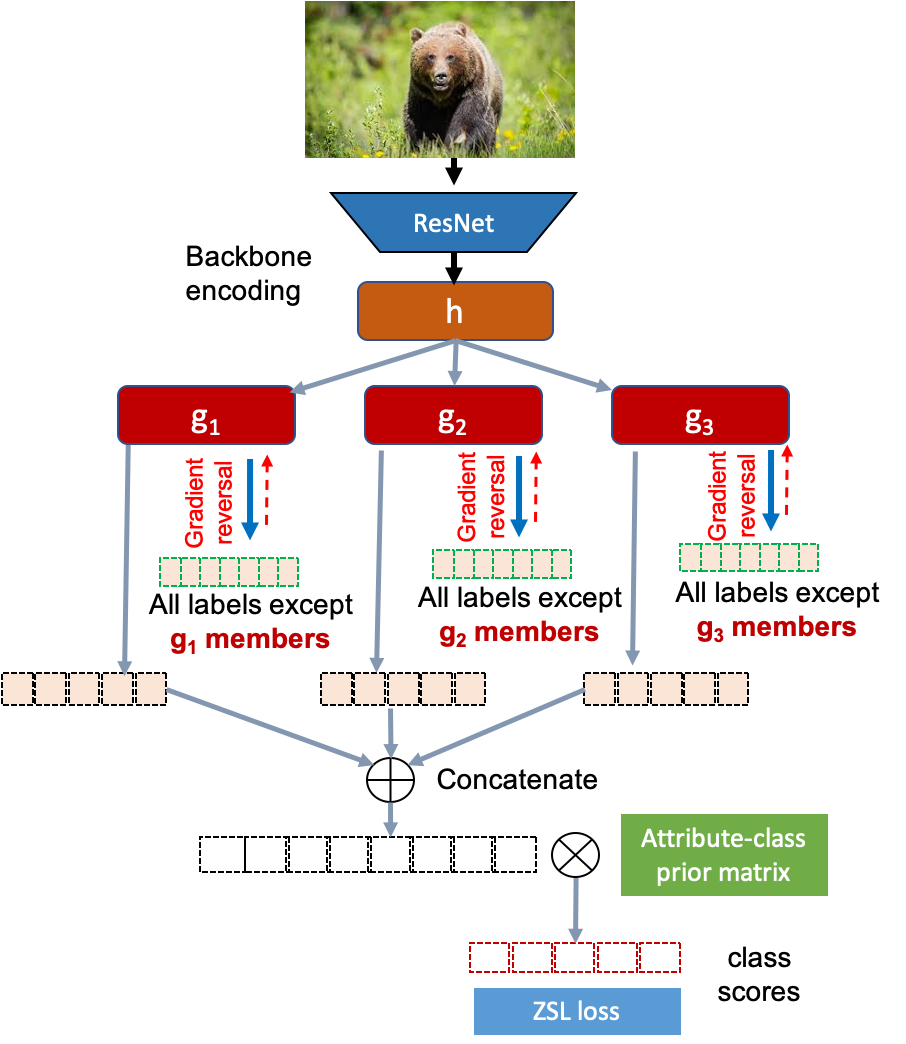}
\caption{Proposed model architecture illustrated for 3 groups of attributes for brevity. Each group ($g_k$) is adversarially trained with \emph{all} remaining groups. The implicit attribute scores and the class-attribute mapping is used to determine the class prediction loss. 
}
\label{fig:proposed_zsl}
\end{figure}

Our proposed model architecture discussed in Section 3.3.3, is illustrated in Figure \ref{fig:proposed_zsl}. Following are additional details to aid reproducibility of the model architecture and training. The complete code base is provided in the supplementary material. 

\begin{itemize}[leftmargin=*]
\item The best number of groups formed by spectral co-clustering (between 3 and 10) is found empirically per dataset and per classifier.
\item For building our proposed \emph{g}AL architecture, we first attach 500 linear layers to the input Res101 features. Next, we add another 100 layers to form the latent group representations. These are fully connected to the primary and adversarial attribute prediction neurons. None of the internal layers use any non-linear activation function. The primary group attribute predictions are concatenated before being used as input to any of the 4 classifiers (ALE, DeViSE, SJE or \texttt{softmax}). The adversarial attribute predictions go through an additional sigmoid activation layer before being used to compute the adversarial group losses (balanced bce loss).
\item All weights in the final classifier layers (both primary and adversarial) are penalized by L2 regularization. The internal linear layers are regularized by Dropout with dropout probabilities between 0.2 to 0.5.
\item All models are optimized using SGD with nesterov momentum of 0.9. Batch size is picked from \{64, 128\} and learning rate from \{0.01, 0.001\}.
\item Adversarial weight $\lambda$ and the margin for SVM-rank based losses (ALE, SJE, DeViSE) are picked from a large parameter sweep for best validation error.
\item We use PyTorch 1.2.0 to implement our algorithms and run all experiments on a single Tesla K80 GPU.
\end{itemize}



\end{document}